\documentclass[runningheads]{llncs}
\usepackage{amsmath,amssymb,amsfonts}
\usepackage{algorithm}
\usepackage{algorithmic}
\usepackage{graphicx}
\usepackage{textcomp}
\usepackage{xcolor}
\usepackage{listings}
\usepackage{tabularx}
\usepackage{url}
\usepackage[numbers,sort]{natbib}
\usepackage{hyperref} 

\usepackage[T1]{fontenc}  
\usepackage{courier}      

\def\BibTeX{{\rm B\kern-.05em{\sc i\kern-.025em b}\kern-.08em
    T\kern-.1667em\lower.7ex\hbox{E}\kern-.125emX}}
\begin{document}

\title{Robo-CSK-Organizer: Commonsense Knowledge to Organize Detected Objects for Multipurpose Robots}
\titlerunning{Robo-CSK-Organizer} 

\author{Rafael Hidalgo\inst{1} \and 
Jesse Parron\inst{1} \and 
Aparna S. Varde\inst{1} \and 
Weitian Wang\inst{1}}

\authorrunning{R. Hidalgo et al.} 

\institute{School of Computing, Montclair State University (MSU), NJ, USA\\
\email{\{hidalgor2,parronj1,vardea,wangw\}@montclair.edu}}

\maketitle


\begin{abstract}
In the rapidly evolving field of robotics, integration of commonsense knowledge (CSK) in AI systems is becoming highly crucial to enhance the decision-making capabilities of robots, especially in next-generation multipurpose environments. This paper presents Robo-CSK-Organizer, a pioneering system that employs CSK, via a classical knowledge base, to facilitate sophisticated task-based object organization helpful in multipurpose robots. Unlike systems relying solely on deep learning tools such as ChatGPT, our Robo-CSK-Organizer system stands out in various crucial aspects. This includes: (1) its ability to resolve ambiguities and maintain consistency in object placement; (2) its adaptability to diverse task-based classifications; and moreover, (3) its contributions to explainable AI (XAI), consequently helping to foster trust and human-robot collaboration. This system’s efficacy is underlined by DETIC (DEtector with Image Classes), an advanced extension of Detectron2 for object identification; BLIP (Bootstrapping Language-Image Pre-training) for context discernment; and most vitally by the adaptation of ConceptNet, a well-grounded commonsense knowledge base for reasoning based on semantic as well as pragmatic knowledge. While we deploy ConceptNet to extract CSK, the process in Robo-CSK-Organizer is generic enough to be replicated with other state-of-the-art knowledge bases. Controlled experiments and real-world applications, synopsized in this paper, make Robo-CSK-Organizer demonstrate superior performance in placing objects in contextually relevant locations, highlighting its clear capacity for commonsense-guided decision-making closer to the thresholds of human cognition. Hence, Robo-CSK-Organizer makes valuable contributions to Robotics and AI.  
\end{abstract}

\keywords{AI-Robotics Bridge, Commonsense Reasoning, Explainable Models, Multipurpose Robots, Next-Generation AI Systems, Task Classification}

\section{Introduction to Robo-CSK-Organizer}

\begin{figure}[htbp]
\centerline{\includegraphics[width=\textwidth] {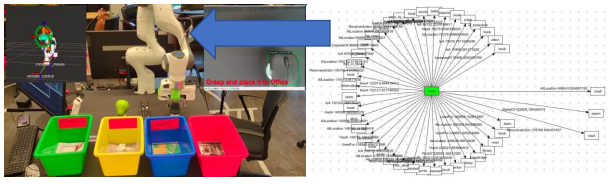}}
\caption{\footnotesize Graphical abstract of Robo-CSK-Organizer
}
\label{fig:abstract}
\end{figure}

The expanding role of multipurpose robots necessitates the development of transparent, intelligent systems capable of handling complex, context-driven operations. Traditional robotic arms, while proficient in assembly tasks, often struggle with nuanced decision-making in dynamic environments. Our work addresses this challenge by introducing Robo-CSK-Organizer (see Fig.~\ref{fig:abstract}), a system that demonstrates Explainable AI (XAI) via Commonsense Knowledge (CSK) to enhance the decision-making capabilities of robots. Imbibing CSK through the ConceptNet knowledge base, this system aims to address the opaque-box issue in AI (typically found in pure deep learning systems), characterized by a lack of decision-making clarity --- a growing concern in the realm of robotics where precision and trust are paramount \cite{zednik2019solving}.

\normalsize

The Robo-CSK-Organizer system is proposed to address the limitations of current AI systems, which often fall short in tasks requiring deep contextual awareness. Traditional AI models may often struggle with differentiating between similar objects in varied contexts, such as distinguishing a child’s toy from a pet’s toy, or understanding the appropriate placement of objects in household settings. Robo-CSK-Organizer bridges this gap by adequately deploying commonsense knowledge (CSK) to enhance the organization of detected objects for task classification in multiple avenues.  

Note that CSK differs from encyclopedic knowledge \cite{nastase2009summarizing}, \cite{joshi2020contextualized} (where AI systems far surpass humans). ``Common sense'' is naturally found in humans who acquire it inherently at birth, enhance it with further growth, and use it for intuitive reasoning. Machines on the other hand are not endowed with CSK by default and hence can often find it challenging to conduct reasoning intuitively unless pre-programmed with rigorous training \cite{davis2015commonsense}. For instance, it is very easy and in fact rather obvious for a human to know that the ``door'' of a \emph{refrigerator} should not remain open (except while placing things in it or taking them out) \cite{choi2022curious}. Conversely, the ``door'' of an \emph{office} can certainly be open and is often ajar - a related fact also quite obvious to humans. Such knowledge is very subtle and is thus considered really simplistic or too common! Yet it can often be crucial in decision-making. Therefore, the role of CSK can be vital in modern-day AI systems as noticed significantly in the literature \cite{zellers2019recognition}, \cite{cambria2022senticnet}. 

Robo-CSK-Organizer precisely addresses this concept of CSK in AI. More specifically, it employs a robust semantic network from a classical knowledge base, namely ConceptNet \cite{speer2018conceptnet}, hence enabling robots to make more well-informed decisions based on contextual cues. This is in line with the logic of harnessing commonsense knowledge to augment machine intelligence \cite{tandon2018commonsense}. It thrives on the adequate extraction and compilation of CSK from a knowledge base, which is a non-trivial task \cite{razniewski2021information}, and can be crucial in AI applications. 

\section{The Robo-CSK-Organizer Approach}

\begin{figure}[htbp]
\centerline{\includegraphics[width=\textwidth] {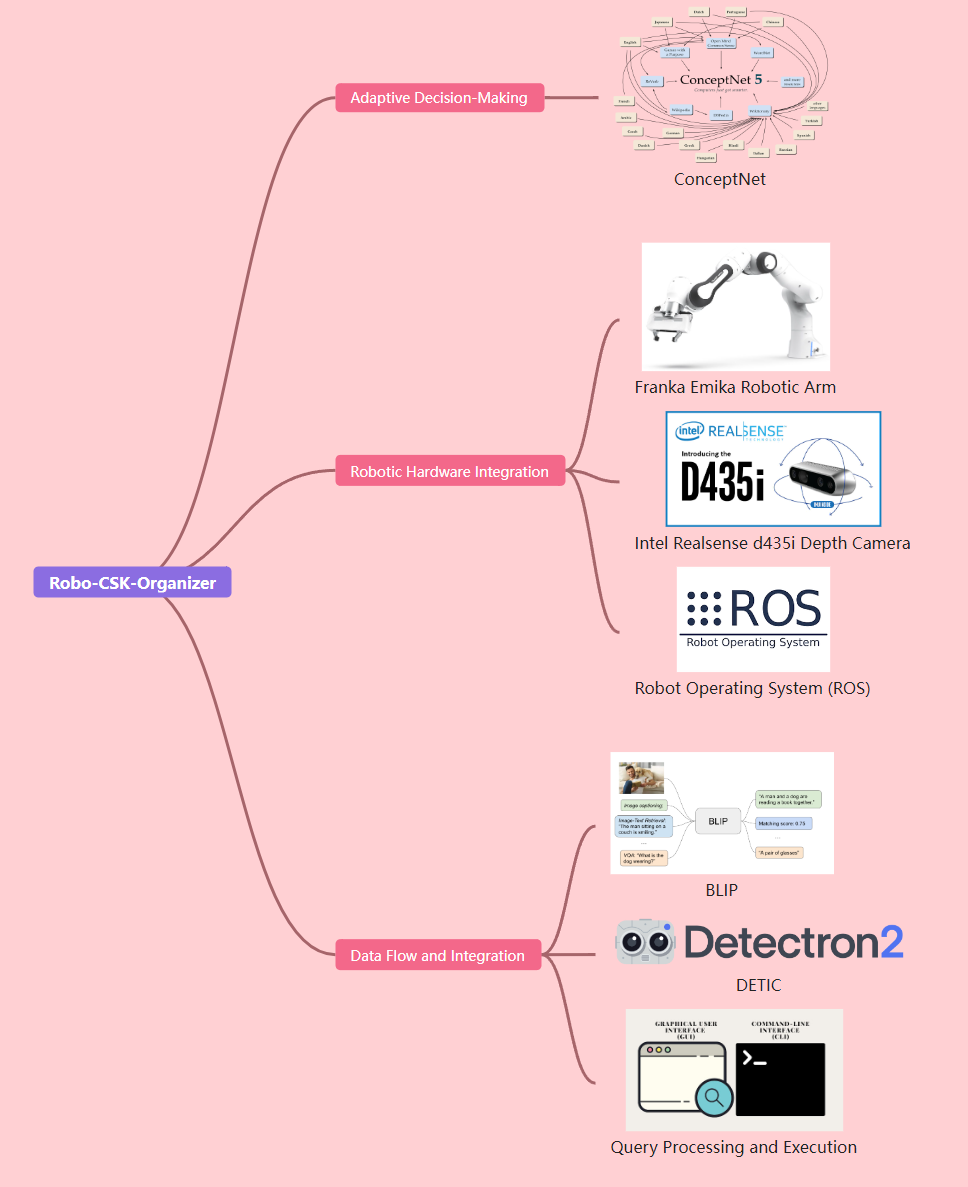}}
\caption{System diagram of Robo-CSK-Organizer}
\label{fig:system_diagram}
\end{figure}

The system diagram of Robo-CSK-Detector, illustrative of its functioning approach, appears in Fig.~\ref{fig:system_diagram} Its salient features are as described below. 

\begin{enumerate}
    \item \textbf{Resolving ambiguity in object categorization:} Classification of objects, such as deciding whether a pear belongs in the kitchen or the garden, highlights the ambiguity inherent in object categorization today. This issue extends to the difficulty of providing comprehensive labeled training examples for every possible object arrangement and is effectively achieved by Robo-CSK-Organizer by harnessing CSK in a task-relevant manner. 
    
    \item \textbf{Maintaining consistency in object placement:} Ensuring consistency in placing objects is crucial, particularly to build trust among users (e.g. helpful in human-robot collaboration) as well as among other robots in a multi-robot environment. Robots must reason with basic commonsense as well as domain knowledge and manage contextual variations to in order to ensure reliable object placement. This is well-achieved by Robo-CSK-Organizer with clear reasoning paths. 
    
    \item \textbf{Depicting task relevance and adaptability:} Robots demonstrating adaptability in prioritizing tasks based on context, e.g. choosing between gardening and culinary activities, is vital. This is especially with respect to probabilistic uncertainties in sensing and navigation. Robo-CSK-Organizer handles this very well due to its systematic approach guided by a well-grounded knowledge base. 
    
    \item \textbf{Fostering explainability in AI systems:} A critical aspect of XAI (Explainable AI) is ensuring that AI systems are not just intelligent but also comprehensible and interpretable. Robo-CSK-Organizer excels in explainability, surpassing other systems (e.g. ChatGPT-based organizers) by adequately harnessing CSK due to which decision-making processes are more transparent and understandable.
\end{enumerate} 

\begin{figure}[htbp]
\centerline{\includegraphics[width=\textwidth] {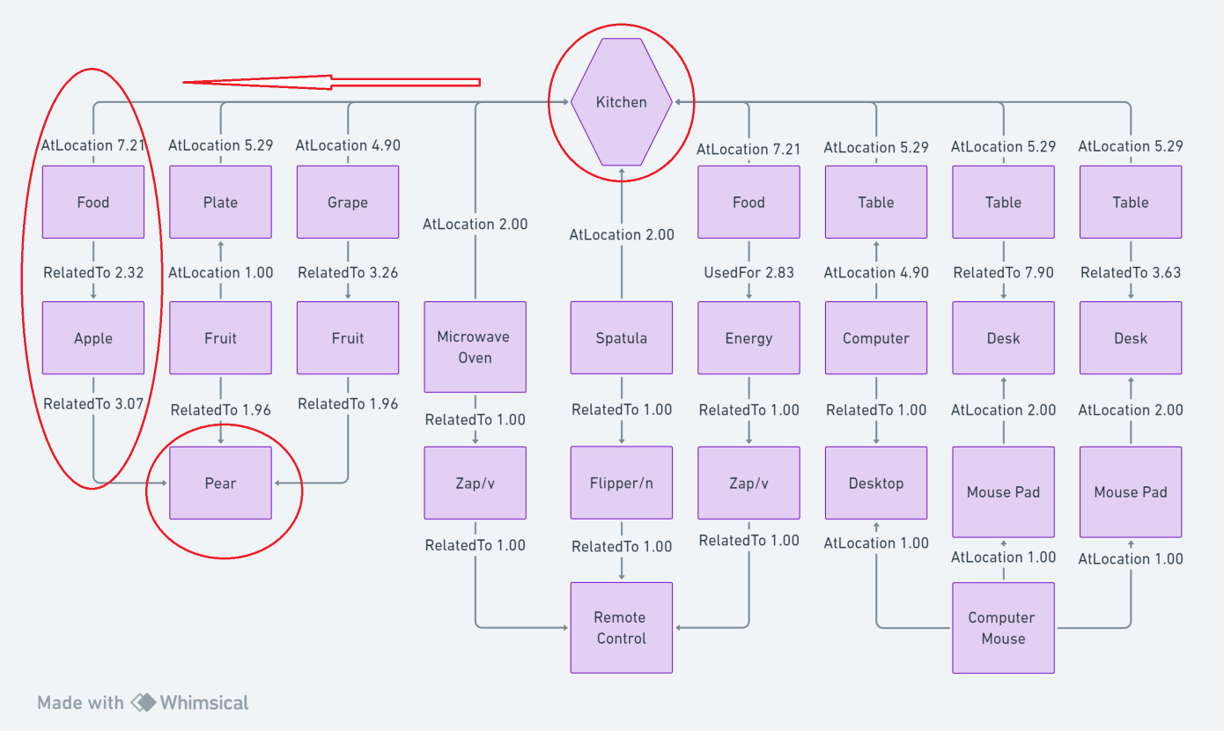}}
\caption{\footnotesize ConceptNet-based decision pathway in Robo-CSK-Organizer. Illustration of the object categorization process for a pear, highlighting the selection of the optimal path based on 'AtLocation' edge weight and relational connections between 'food', 'apple', and 'pear'}
\label{fig:conceptnet_example}
\end{figure}

In Fig.~\ref{fig:conceptnet_example}, Robo-CSK-Organizer's utilization of ConceptNet for object categorization in a kitchen setting is visually depicted. This figure illustrates the decision-making pathway for categorizing a pear, among other items. Specifically, it shows 3 potential paths from ``kitchen'' to ``pear''. The system selects the path with the highest ``AtLocation'' edge weight (in this case, 7.21), indicating the common location of food in a kitchen. This path is further delineated by linking ``apple'' to ``food'' (via a ``RelatedTo'' edge) and subsequently connecting ``apple'' to ``pear''. This logical sequence leads Robo-CSK-Detector to place the pear in the kitchen, exemplifying the system's reasoning process.

\normalsize 

Designed with modules for object detection, context recognition, and semantic analysis, Robo-CSK-Organizer not only classifies objects but also interprets their appropriate placement within various contexts. This approach significantly enhances the transparency and interpretability of AI decisions, addressing the critical challenge of explainability in AI systems. It can thus be considered analogous to systems that delve into aspects such as spatial commonsense \cite{garg2020guessing}, \cite{GargTV22}, \cite{persaud2017enhancing} for object recognition and training of autonomous systems. For instance, if there is an error, its traceability is highly facilitated in Robo-CSK-organizer (versus deep learning based systems, e.g, those using ChatGPT for training). Hence, Robo-CSK-Organizer can help robots learn from their mistakes and correct themselves, thus getting better in their performance. Moreover, the XAI contribution of Robo-CSK-Organizer is helpful when humans and robots work together, i.e. for human-robot collaboration, as humans are able to understand the actions of the robots much better, along with the reasons behind the robots' decisions. The same logic applies to numerous robots working together. This enhances trust in the realm of robotics. All these facets are vital, especially with the growing prevalence of multipurpose robots, heading towards next-generation advancements. 

The main functioning of Robo-CSK-Organizer, focusing on its reasoning, is outlined in Algorithm~\ref{alg:robo-csk-organizer} here. 

\begin{algorithm}
\caption{Robo-CSK-Organizer Reasoning}
\label{alg:robo-csk-organizer}
\begin{algorithmic}[1]
\REQUIRE $\mathcal{V}$ (Video feed), $\mathcal{C}$ (ConceptNet knowledge base)
\ENSURE $\mathcal{O}_{\text{sorted}}$ (Objects sorted into appropriate contexts)

\STATE $\text{Initialize robot vision system } \mathcal{R} \leftarrow \text{Detectron2}$
\STATE $\text{Scan context bins using } \mathcal{B} \leftarrow \text{BLIP, store context in } \mathcal{S}_{\text{csv}}$
\FOR{$f \in \mathcal{V}$}
    \STATE $\mathcal{D} \leftarrow \text{Detect and label objects in } f \text{ using } \mathcal{R}$
    \FOR{$o \in \mathcal{D}$}
        \STATE $\mathcal{K}_o \leftarrow \text{Query } \mathcal{C} \text{ for context of } o$
        \IF{$\mathcal{K}_o \text{ matches context in } \mathcal{B}$}
            \STATE $\text{Place } o \text{ in matched context bin}$
        \ELSE
            \STATE $\text{Continue to next object}$
        \ENDIF
    \ENDFOR
\ENDFOR
\STATE $\text{Optional: Display annotated frame from } f$
\end{algorithmic}
\end{algorithm}

This algorithm that highlights the main functioning of Robo-CSK-Organizer operates on 2 primary inputs: a video feed $\mathcal{V}$ and the ConceptNet knowledge base $\mathcal{C}$. Its goal is to sort detected objects into their appropriate contexts. Initially, the robot's vision system $\mathcal{R}$, implemented using Detectron2, is initialized. Thereafter, the context bins are scanned and recognized using BLIP ($\mathcal{B}$), with the contexts stored in a CSV (comma-separated variable) format ($\mathcal{S}_{\text{csv}}$). For each frame $f$ in the video feed $\mathcal{V}$, objects are detected and labeled as $\mathcal{D}$. Each detected object $o$ is then checked against $\mathcal{C}$ to determine its context $\mathcal{K}_o$. If this context matches that of a bin in $\mathcal{B}$, the object is placed in the corresponding bin. The process continues for each object in the frame, and the annotated frame can be displayed as needed. 

This algorithm is implemented into the Robo-CSK-Organizer system using Python and is integrated with a robotic arm in our laboratory, namely, the CRoSS Lab (Collaborative Robotics and Smart Systems Lab at our university). Robo-CSK-Organizer is then executed using various real-world objects. Details of its execution are mentioned next in the respective parts of its system demonstration.

\section{System Demo and Evaluation}

In order to demonstrate the efficacy of Robo-CSK-Organizer, it is compared with a baseline task organizer that uses the well-known ChatGPT for guidance. We thus present the following. 

\textbf{Object Detection:} Both the systems, our Robo-CSK-Organizer and the ChatGPT baseline, use DETIC (”DEtector with Image) ensure a broad evaluation spectrum. Specific context groups, particularly domestic locations (e.g. kitchen, garden, pantry, dining room) are chosen for evaluation. These contexts are relevant to the selected object categories and provided with a controlled environment for testing. An advanced extension of Detectron2 for object detection, which contains over 21,000 classes  \cite{Wu2019Detectron2} \cite{zhou2022detecting} is used here to provide choices of classes for object organization. Each object is queried against each system (Robo-CSK-Organizer / ChatGPT) 10 times, asking it to organize the object into one of the provided contexts. Responses from both systems are recorded for each iteration, and the most frequent context is identified as the predominant choice for object placement, 

\textbf{Context Recognition:} BLIP (Bootstrapping Language-Image Pre-training) is employed to identify contexts such as the kitchen or office, and thus generate room captions for enhanced clarity \cite{li2022blip}. Note that the usage of such software can be helpful in a variety of applications, e.g. image personalization via text by harnessing diffusion models \cite{hidalgo2023personalizing}. The hardware foundation for both systems (i.e. Robo-CSK-Organizer and the ChatGPT baseline) includes a Franka Emika robotic arm \cite{HaddadinFrankaEmika} with an Intel Realsense D435i camera \cite{IntelRealSense2023}, integrated with ROS. Robo-CSK-Organizer works with this hardware, and incorporates CSK-based reasoning (See Algorithm~\ref{alg:robo-csk-organizer}). This is the key to addressing the opaque-box issue in AI, aiming for clearer and more transparent object sorting. 

Note that the pivotal distinction between the 2 systems lies in the functioning approach for sorting objects into relevant contexts. While ChatGPT relies solely on prior training with deep learning, Robo-CSK-Organizer applies commonsense knowledge due to which it can be more adept in successfully handling first-time scenarios as well. More details appear next. 

\textbf{Robo-CSK-organizer:} It harnesses a classical knowledge base called ConceptNet \cite{speer2018conceptnet} for semantic insights and commonsense reasoning. It infers object locations using metrics known as \emph{edge weight} and \emph{degree of separation}, prioritizing paths based on these factors. While ConceptNet is chosen for its user-friendly interface and clear path logic, we claim that other relevant CSK knowledge bases can also be used.

\textbf{ChatGPT baseline:} A ChatGPT-trained organizer is used as a baseline; it relies on generative pre-trained transformer models for its decision-making, processing text-based inputs to infer object locations and categorizations. This approach, though adept in language processing, does not integrate a structured commonsense knowledge base. Consequently, the ChatGPT-based organizer’s decisions are more influenced by pre-trained patterns in textual data rather than explicit semantic relationships and intuitive logical reasoning. This can affect consistency and transparency in decision-making in complex or ambiguous scenarios, notably (but not limited to) those encountered for the first time. 

In our comprehensive evaluation, we conduct experiments to assess the performance of Robo-CSK-Organizer and the Chat-GPT baseline across various contexts. The key aspects of these experiments are focused on ambiguity resolution, consistency, task-relevance adaptability, and explainability. A summary of our exhaustive experimentation is presented below. 

\subsection{Ambiguity Resolution}
\begin{figure}[htbp]
\centerline{\includegraphics[width=\textwidth]{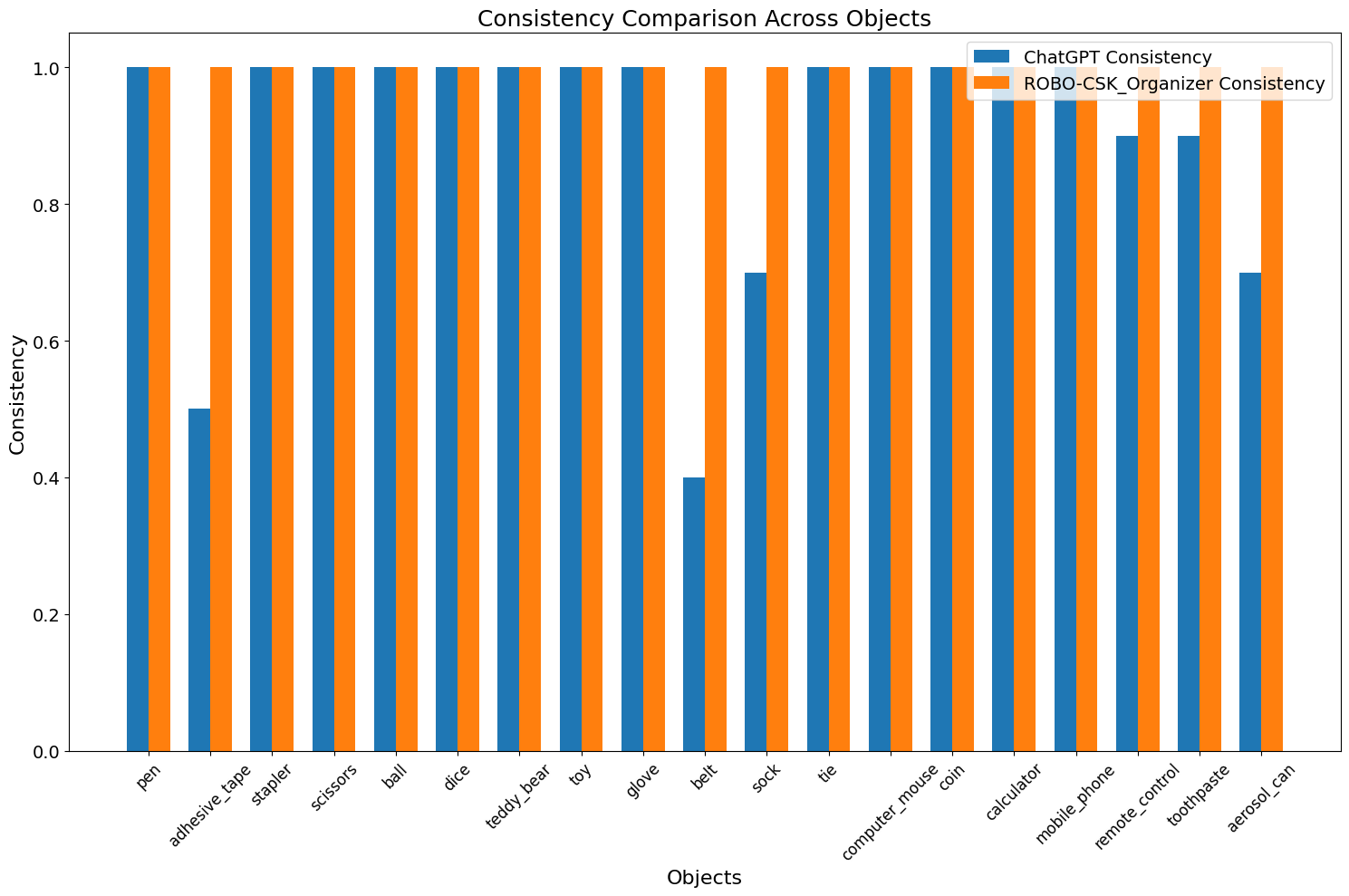}}
\caption{\footnotesize Robo-CSK-Organizer has 100\% consistency in all object-location pairs. ChatGPT baseline is not as consistent for all objects, specifically for adhesive tape, belt, sock, remote control, toothpaste, and aerosol can}
\label{fig:consistency}
\end{figure}

Both systems are tested on their ability to resolve ambiguous contexts using a variety of objects. Robo-CSK-Organizer as well as ChatGPT baseline performances are evaluated against a ground truth established by semantic similarity scores from state-of-the-art paradigms such as FastText, Word2Vec, and GloVe models that can be widely accepted as gold standards. The results (See Fig.~\ref{fig:consistency}), show that Robo-CSK-Organizer has notable accuracy. 


\subsection{Ensuring Consistency}
The consistency experiments aim to evaluate the stability and repeatability of Robo-CSK-Organizer versus the ChatGPT baseline when faced with identical queries across multiple iterations. This measure of consistency is vital for reliable knowledge organization systems, as it reflects the systems’ ability to consistently choose the same context for an object through numerous trials. In the methodology, objects from various categories (e.g. personal items, clothing, office supplies, and toys) are selected.  Robo-CSK-Organizer achieves 100\% consistency rate across all object-location pairs; this can be attributed to the static nature of the ConceptNet knowledge graphs that it utilizes in its decision-making. In contrast, the ChatGPT Organizer displays less consistency, particularly for objects such as adhesive tape, belt, sock, remote control, toothpaste, and aerosol can; possibly indicating that pre-training alone may not always yield consistent results in systematic object organization. 

\normalsize

\subsection{Task-Relevance Adaptability}
These experiments evaluate the systems’ ability to adapt their responses to different context-specific directives. The objective is to assess whether Robo-CSK-Organizer and the ChatGPT baseline can re-calibrate their responses when directed to focus on alternative contexts differing from their initial preferences. The experiments commence with an initial straightforward assessment, querying ``apple'' against 4 contexts (kitchen, living room, bedroom, bathroom) without any focused directives. This determines the systems’ natural inclination or preference for context association. In the adaptability testing phase, the systems are prompted to focus on the remaining 3 contexts, one at a time, to observe if they can adapt their responses when a specific context is emphasized. This is analogous to humans adapting to different contexts in real life, e.g. if you specifically tell a human not to place apples in a kitchen (for any reason, e.g. the kitchen is too small or it is being cleaned for pest control). then the human should intuitively find another good place for the apples rather than placing them in the kitchen again. Accordingly, it is interesting to assess how robotic systems would behave in such situations. 

Likewise, data for the initial as well adaptability tests are collected by repeating each context-query 10 times. Responses are compiled into respective data frames for detailed analysis. The focused contexts for the adaptability tests are based on preference, with the most preferred context excluded to emphasize the remaining contexts. The initial phase identifies kitchen as the clear preference for sorting apples for both the ChatGPT-based organizer and Robo-CSK-Organizer. In the adaptability phase, there are observable shifts in Robo-CSK-Organizer’s response, as desired, i.e. it is more adaptable when needed. The paths Robo-CSK-Organizer employs for sorting, leading to object placement, are as follows:

\begin{itemize}
    \item Path: Kitchen (AtLocation) $<$- food (RelatedTo) $>$ apple
    \item Path: Bedroom (AtLocation) $<$- house (AtLocation) $<$- apple
\end{itemize}

\begin{figure}[htbp]
\centerline{\includegraphics[width=\textwidth]{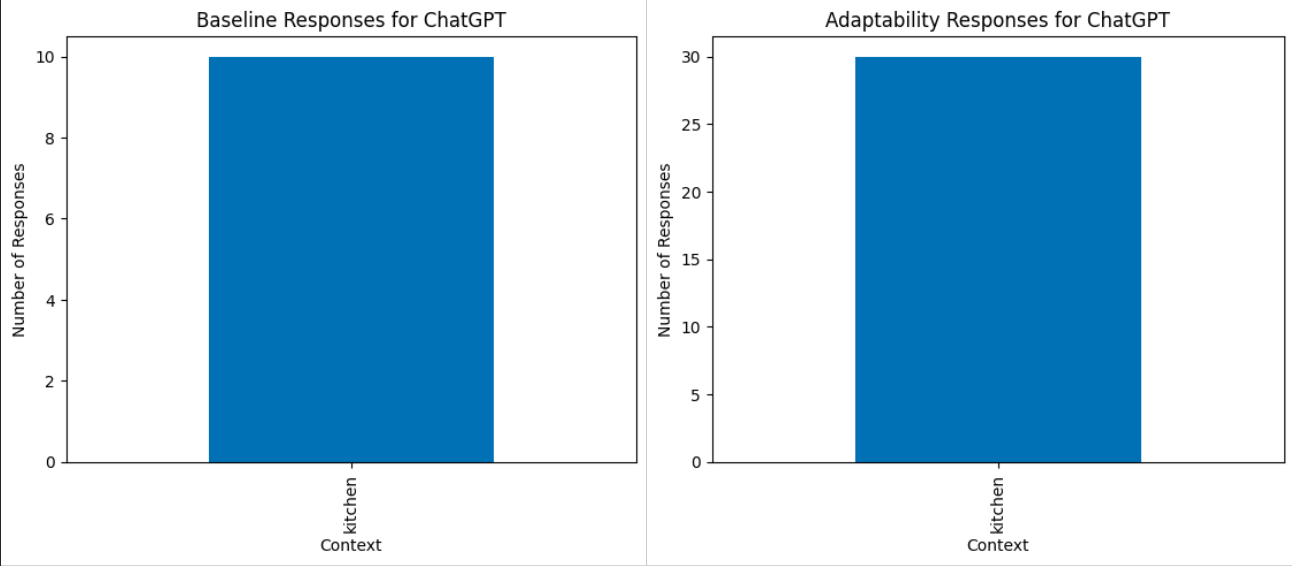}}
\caption{\footnotesize In the adaptability phase, ChatGPT’s implementation does not cause any observable shifts, despite requesting it to change its context.}
\label{fig:adaptability1}
\end{figure}

\begin{figure}[htbp]
\centerline{\includegraphics[width=\textwidth] {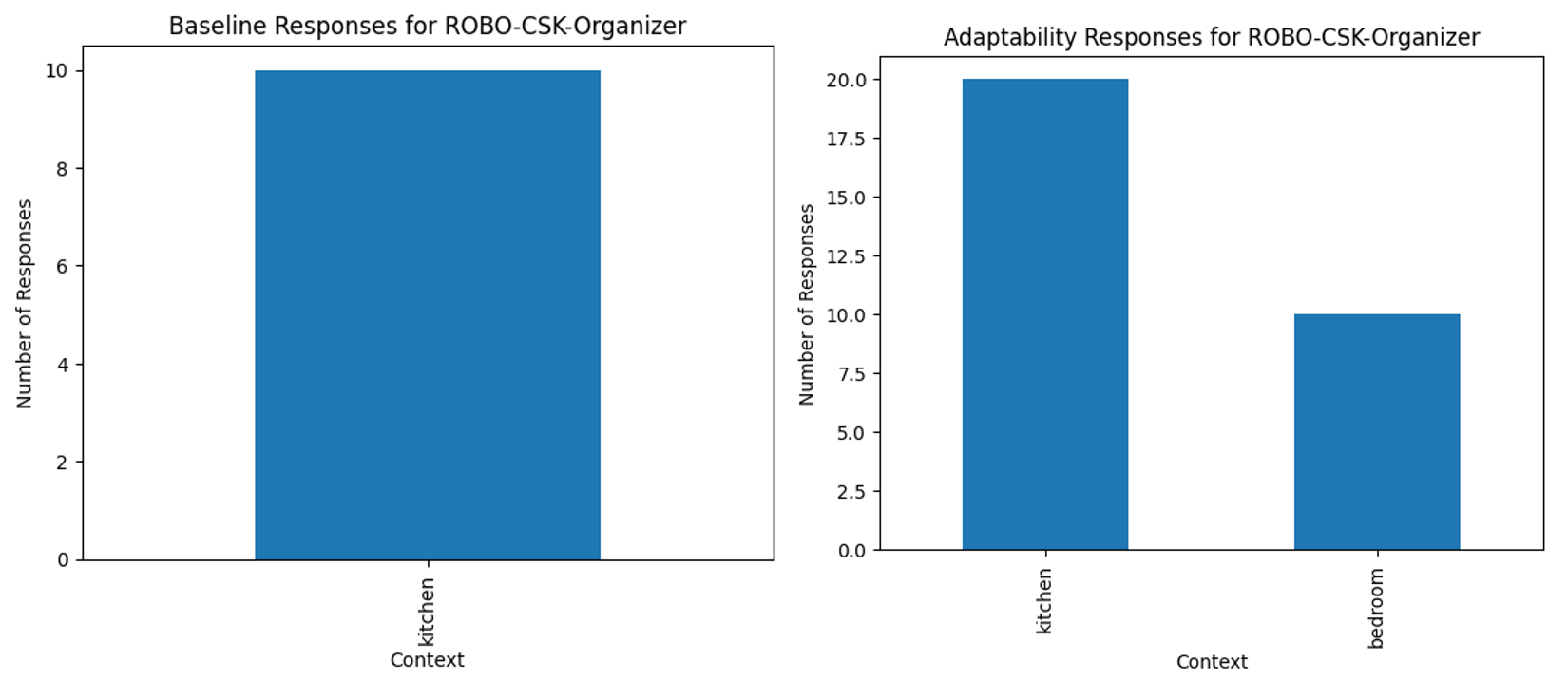}}
\caption{\footnotesize There are observable shifts on the Robo-CSK-Organizer in one of the locations, after requesting context-switching.}
\label{fig:adaptability2}
\end{figure}

Figs~\ref{fig:adaptability1} and~\ref{fig:adaptability2} here provide a well-summarized visual representation of these findings.

\normalsize

\normalsize

\subsection{Explainability}

The experiment on explainability assesses the Robo-CSK-Organizer system and the ChatGPT-based system as per their abilities to elucidate their decision-making processes. This aspect is crucial for building user trust and understanding, considering the fact that there are situations where decisions may seem counter-intuitive at times. Robo-CSK-Organizer utilizes Detic for object detection, BLIP for context recognition, and ConceptNet for commonsense knowledge. It can provide logical paths for its decisions, enhancing transparency. For instance, during its incorrect placement of beer into the playroom, Robo-CSK-Organizer provides a clear logical path: \emph{playroom (UsedFor) fun (RelatedTo) party (RelatedTo) beer}. This path demonstrates the connection of concepts leading to the system’s conclusion. In contrast, the ChatGPT-based organizer, relying basically on deep learning models, functions as an opaque-box. It is unable to ascertain the explicit reasoning behind its decisions, e.g. placing ``scissors'' into a ``playroom'' (which can be a potentially hazardous decision). Hence, the ChatGPT baseline is lacking in a clear explainable framework. This can pose problems in error-correction, thus adversely impacting performance. 

This distinction highlights that while both systems may err, analogous to the adage ``\emph{to err is human}'', Robo-CSK-Organizer’s explainability allows for better understanding and correction of these errors. Explainability is essential, especially in robotic systems where precision and safety are critical, contributing to user trust and understanding of AI decisions. Note that such explainability can in turn help in explicit communication with various AI systems, including intelligent agents in mobile apps, e.g. it can help to enhance existing apps with virtual voice agents \cite{kalvakurthi2023hey} by adding more image-based functions where adequate object recognition is crucial. Hence, it can indirectly help a different type of robot, including a chatbot or a virtual voice assistant. All these systems would benefit from easier comprehension and enhanced interpretability. Hence, explainable AI plays a vital role here. 

\begin{figure}[htbp]
\centerline{\includegraphics[width=\textwidth] {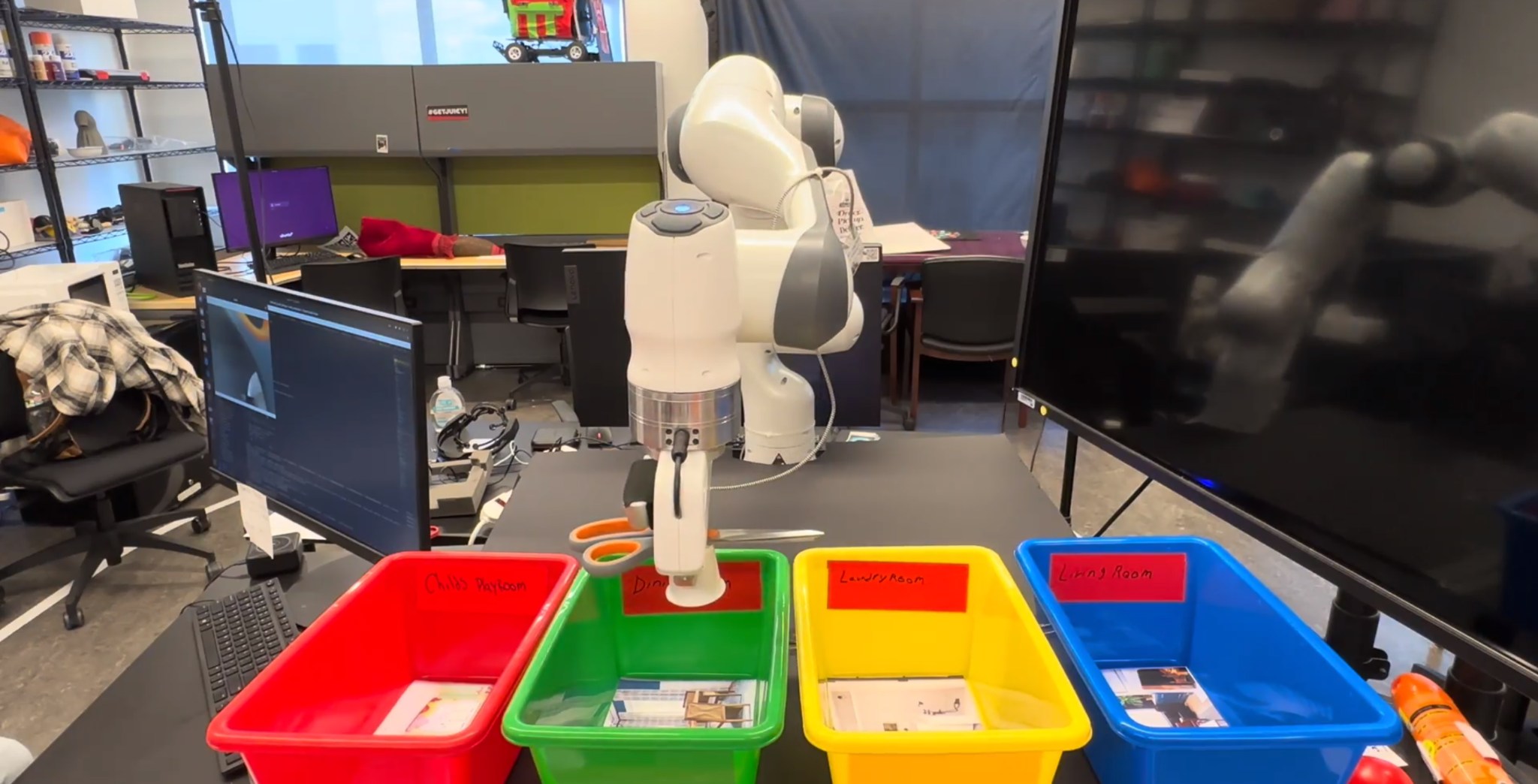}}
\caption{\footnotesize Robo-CSK-Organizer's (quite adequate) placement of scissors into the dining room}
\label{fig:explain1}
\end{figure}

\begin{figure}[htbp]
\centerline{\includegraphics[width=\textwidth] {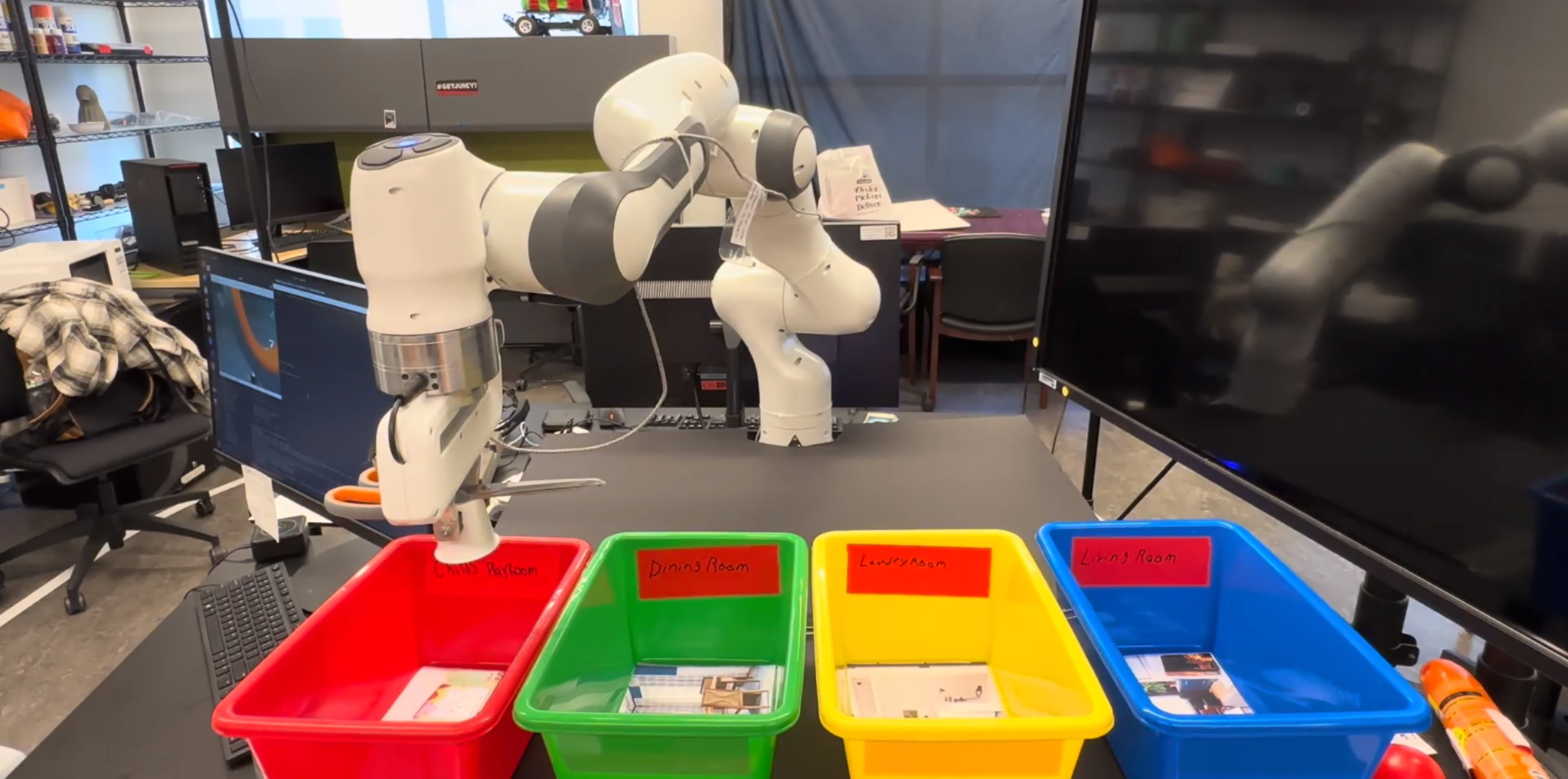}}
\caption{\footnotesize ChatGPT-Organizer's (rather dangerous) placement of scissors into the child’s playroom}
\label{fig:explain2}
\end{figure}

Focusing on such aspects, Figs~\ref{fig:explain1} and~\ref{fig:explain2} illustrate how Robo-CSK-Organizer and the ChatGPT-based organizer derive their decisions in the placement of ``scissors''. The comparative analysis emphasizes the Robo-CSK-Organizer’s strengths in consistency, adaptability, and explainability. The findings of our experimentation thus underscore the importance of integrating structured knowledge bases in AI systems. This fact is highlighted here, considering various scenarios for domestic environments. 

\normalsize

\normalsize

Finally, we synopsize the comparative evaluation of Robo-CSK-Organizer and the ChatGPT baseline in a TABLE 1 here. 

\begin{table}[h]
\centering
\caption{Comparison of Robo-CSK-Organizer and ChatGPT}
\label{tab:comparison}
\resizebox{\textwidth}{!}{%
\begin{tabular}{|l|l|l|}
\hline
\textbf{Parameter} & \textbf{Robo-CSK-Organizer} & \textbf{ChatGPT} \\ \hline
Approach & Uses ConceptNet, logical paths & Generative transformer models \\ \hline
Ambiguity Resolution & Noticeable Accuracy & Varies based on trained patterns \\ \hline
Consistency & 100\% consistent across all classes & Consistency varies depending on class \\ \hline
Task-Relevance Adaptability & Adaptable to directives & Limited adaptability shown \\ \hline
Explainability & High; clear paths & Lower; opaque due to AI \\ \hline
Decision-Making Basis & Semantic, pragmatic CSK & Textual data patterns \\ \hline
Hardware Integration & Robotic systems integration & Robotic systems integration \\ \hline
Use of AI & AI with knowledge bases & Primarily AI-driven \\ \hline
\end{tabular}%
}
\end{table}

\section{Discussion on Main Contributions}

This research presents significant contributions to the field of AI and robotics, particularly in the development and application of commonsense knowledge for task classification in multipurpose robots. Robo-CSK-Organizer, as a pioneering system, stands out in several key areas when compared to existing systems (e.g. a baseline organizer using ChatGPT). 

\subsection{Novel Integration of CSK}
Robo-CSK-Organizer’s integration of CSK through ConceptNet is a major advancement. Unlike systems that rely primarily on deep learning, Robo-CSK-Organizer utilizes structured knowledge bases in addition to machine learning, to enhance decision-making transparency and accuracy. This integration allows the system not only to recognize and categorize objects but also to self-understand the task contexts, leading to more intuitive and contextually appropriate object operations.

\subsection{Superiority in Consistency and Adaptability}
Our experiments depict Robo-CSK-Organizer’s superiority in consistency and task-relevance adaptability. It achieves a 100\% consistency rate, which is crucial for user trust and predictability. Its ability to adapt to different context-specific directives showcases its potential in dynamic and changing environments, which is essential for practical applications such as next-generation multipurpose robots, e.g. those meant to be helpful in domestic settings.  

\subsection{Advancements in Explainability}
Robo-CSK-Organizer makes strides in explainability, a key aspect of XAI (Explainable AI). Its ability to provide logical paths for its decisions enhances user understanding and trust, particularly in situations where decisions might seem counter-intuitive. This feature sets it apart from more opaque-box systems such as those relying solely on deep learning. The more explainable a system is, the easier it is to work with, especially when multiple robots work together and / or humans and robots collaborate with each other. 

\section{Related Work}

The integration of AI and robotics, particularly in domestic environments, has seen a variety of innovative approaches. These methodologies have significantly contributed to the field by enhancing robots' decision-making processes, adaptability, and interaction with their environment.

One approach focuses on using ConceptNet and Google search data for object categorization in domestic robotics, particularly for tidy-up services. This method effectively groups objects into functional categories, thereby aiding robots in more intuitive object handling \cite{6705507}. Another study explores the use of large language models (LLMs) like GPT-3.5 as a repository of CSK for task planning. This demonstrates the potential of language models in enriching the robotic decision-making process \cite{zhao2023large}. Furthermore, paradigms based on the classical neural networks have been adapted to many contexts, ranging from machine translation \cite{CoralloLRSVW22} in text with recurrent neural networks (RNNs) to object recognition in multifaceted scenarios with computer vision models, e.g. VGG-16 \cite{simonyan2015deep} and ResNet-101 \cite{HeZRS15}. The issue of extracting cultural commonsense knowledge and its usefulness in enhancing chatbots has been addressed through a novel approach called CANDLE \cite{NguyenRVW23} with interesting real-world impacts.  

Advances in visual commonsense reasoning introduce the R2C engine \cite{zellers2019recognition} to enhance object recognition, anchoring natural language descriptions in visual data. CSK-Detector \cite{ChernyavskyCSKDetectctor} is an innovative system for object detection in domestic robotics, leveraging CSK from the Dice knowledge base \cite{Chalier2020DiceAJ}; it reduces the need for extensive image annotation. 

The incorporation of CSK from the OMICS database using Description Logic has also been discussed. This integration enables robots to perform more nuanced tasks, showcasing the potential for more context-aware robotics \cite{10.1007/978-3-642-16111-7_17}. Furthermore, the application of CSK in human-robot collaborative tasks has been highlighted, especially in robot action planning for assembly tasks, emphasizing the enhancement of cooperative interactions \cite{Conti2020RobotAP}. Its mathematical modeling insights along with core applications in smart manufacturing have been elaborated \cite{9743276} as well, emphasizing the crucial role of commonsense reasoning.  

Additionally, semantic task planning for service robots in dynamic, open-world environments has been explored. This method leverages natural language understanding and semantic reasoning, addressing the challenges posed by ever-changing environments \cite{fi13020049}. The combination of non-monotonic logical reasoning and incomplete CSK with inductive learning to guide deep learning in robotics is another innovative approach. This integration offers a unique perspective on the convergence of CSK and advanced learning techniques \cite{10.1007/s10458-022-09584-4}. Much of this work builds upon semantic advances over the years \cite{suchanek2011hidden}, \cite{varde2004apriori} that help in managing knowledge and conducting predictive analysis.   

These diverse methodologies underscore the importance of CSK in improving the functionality and intelligence of robotic systems, especially in domestic settings. They have advanced the field by demonstrating effectiveness in task planning, human-robot interaction, and environmental adaptation. 

Building on these foundations, the Robo-CSK-Organizer system represents a significant advancement in the practical application of CSK in robotics. Unlike the existing systems, Robo-CSK-Organizer harnesses CSK in real-world settings for object organization in task-based classification, which is particularly beneficial for multipurpose robots. Its ability to resolve ambiguities and maintain consistency in object placement, adaptability to diverse task classifications, and contributions to explainable AI (XAI) set it apart from the current methodologies. This system not only categorizes and understands objects in various contexts but also intelligently organizes them, demonstrating a novel and practical application of CSK in enhancing the efficiency and functionality of multipurpose robotic systems.

\section{Conclusions and Roadmap}

Robo-CSK-Organizer is a system proposed in this paper that effectively demonstrates XAI via CSK for object organization, mitigating the challenges of non-transparent AI. Our early experiments open up areas for enhancement, e.g. decision paths from a CSK source such as ConceptNet. For instance, misplacement of high heels in the kitchen can be due to semantic overlap with stiletto as heel / knife. When this is clearly explained, potential improvements can be made. In our work, these can include refining relationships (e.g. RelatedTo) so as to provide better contextual accuracy, and using average weights in the knowledge base, not just the weight of the first edge in order to provide more robustness. Additionally, we are committed to refining the algorithmic logic used to identify the most optimal paths based on CSK.

Impacts of Robo-CSK-Organizer are highlighted here.
\begin{enumerate}
    \item Puts forth our objective to quantify enhancements that CSK brings to the reliability and transparency of AI
    \item Elevates the efficacy of robotic decision-making to bring it closer to human cognition
    \item Fosters a broader academic dialogue on commonsense in robots for better interpretation, trust, and explainability
    \item Can be useful in next-generation multipurpose robots, and in human-robot collaboration due to higher clarity.
    \item Can lead to energy savings due to more efficient learning, thus positively impacting sustainable AI.
    \item Well-mounted on an AI-robotics bridge, particularly that of explainable AI and multipurpose robotics.
\end{enumerate}
This paper thus offers our modest contributions to both AI and robotics. We anticipate fruitful, long-lasting impacts.

\section*{Acknowledgments}
\footnotesize
The commonsense knowledge project thrived on a research visit by Dr. Aparna Varde at the Max Planck Institute for Informatics, Saarbrucken, Germany, with further work at Montclair. This work is supported in part by the National Science Foundation under Grants CMMI-2138351 and CNS-2117308. Our experiments are conducted in the CRoSS (Collaborative Robotics and Smart Systems) Lab at Montclair, of which Dr. Weitian Wang is the Director. We thank CESAC (Clean Energy and Sustainability Analytics Center) at Montclair, of which Dr. Aparna Varde is an Associate Director. 

\bibliographystyle{splncs04}
\bibliography{references} 

\end{document}